\documentclass{article}

\usepackage{arxiv}
\usepackage{natbib}
\setcitestyle{square,sort,comma,numbers}
\usepackage[utf8]{inputenc} 
\usepackage[T1]{fontenc}    
\usepackage{hyperref}       
\usepackage{url}            
\usepackage{booktabs}       
\usepackage{amsfonts}       
\usepackage{nicefrac}       
\usepackage{microtype}      
\usepackage{lipsum}
\usepackage{framed,multirow}
\usepackage{booktabs}
\usepackage{algorithm}
\usepackage{algorithmic}%
\usepackage{bbm}
\usepackage{lineno}
\usepackage{amssymb}
\usepackage{latexsym}
\usepackage{amsmath}
\usepackage{subfigure}
\usepackage{placeins}
\usepackage{hyperref}
\usepackage{float}
\usepackage{pdfcomment}
\usepackage{graphics}
\usepackage{graphicx}
\usepackage{xcolor}
\definecolor{newcolor}{rgb}{.8,.349,.1}
\definecolor{newcolor}{rgb}{.8,.349,.1}
\definecolor{mygreen}{rgb}{0.1, 0.75, 0.15}

\newcommand{\tabitem}{~~\llap{\textbullet}~~}

\newcommand{\removeGG}[1]{}

\newcommand{\removeRH}[1]{}

\title{Forensic shoe-print identification: a brief survey}

\author{
  Imad Rida  \\
 Laboratoire BMBI Compiègne \\ 
 Université de Technologie de Compiègne \\ 
  Compiègne, France\\
   \And
 Lunke Fei\\
  School of Computer Science and Technology \\ 
  Guangdong University of Technology \\ 
  47870 Guangzhou, Guangdong China 510006\\
   \AND
  Hugo Proen\c{c}a \\
  Instituto de Telecomunica\c{c}\~oes \\ 
  Department of Computer Science \\
  University of Beira Interior, Covilh\~a \\
   \And
  Amine Nait-Ali \\
  Université Paris-Est, LISSI, UPEC  \\
  94400 Vitry sur Seine , France \\
  \And
  Abdenour Hadid \\
  Center for Machine Vision and Signal Analysis \\ 
  University of Oulu, Oulu, Finland \\
}

\begin{document}
\maketitle

\begin{abstract}
As an advanced research topic in forensics science, automatic shoe-print identification has been extensively studied in the last two decades, since shoe marks are the  clues most frequently left in a crime scene. Hence, these impressions provide a pertinent evidence for the proper progress of investigations in order to identify the potential criminals. The main goal of this survey is  to provide a cohesive overview of the research carried out in forensic shoe-print identification and its basic background. Apart defining the problem and describing the  phases that typically compose the processing chain of shoe-print identification, we provide a summary/comparison  of the state-of-the-art approaches, in order to guide the neophyte and help to advance the research topic. This is done through introducing simple and basic taxonomies as well as summaries of the  state-of-the-art performance. Lastly, we discuss the current open problems and challenges in this research topic, point out for promising directions in this field.
\end{abstract}


The place where the criminals commit their unlawful act namely Scene of Crime (SoC) is for extreme importance for police \cite{huynh2003automatic}. According to Locard's exchange assumption, perpetrator of a crime will inevitably leave something into the SoC \cite{locard1930analysis}. Hence, based on this theory, finding and recovering the physical evidence is crucial and fundamental task in order to identify the criminals and exculpate the unduly accused \cite{vagavc2017detection}.

Fingerprint, blood and hair are examples of clues that can be found in the SoC \cite{liu2017multi,benecke1997dna,robertson2002forensic,buckleton2016forensic,robertson2016interpreting,rida2015improved}. Unfortunately criminals often try to adopt some techniques such as wearing gloves in order to neutralize these clues. On the other hand, although the shoe-prints are not unique, it has been noted that they have greater chance to be present in the SoC than latent fingerprints for instance \cite{thompson2006forensic,bodziak2017footwear}. 

\begin{figure}[!h]
\centering
\subfigure[shoe-print database] {\label{fig:roi:a}\includegraphics[width=8cm]{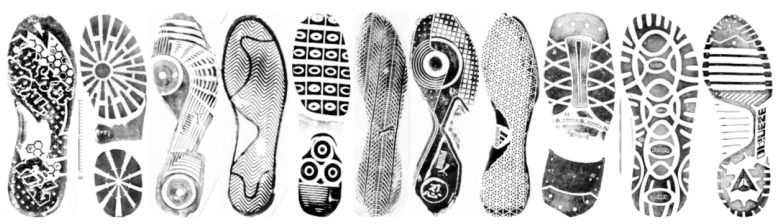}}\\
\subfigure[SoC prints]{\label{fig:roi:c}\includegraphics[width=8cm]{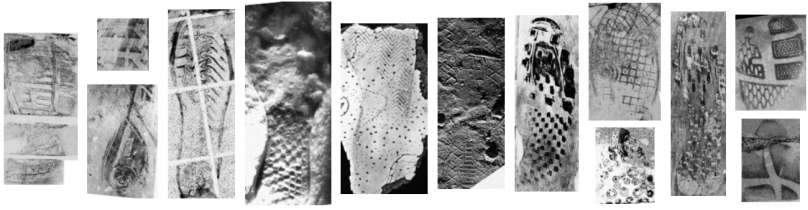}}
\caption{Example of shoe-print images from database and SoC \cite{kong2017cross}.}
\label{fig:roi}
\end{figure}
 


A shoe mark occurs due to the contact of a shoe with a surface (see Figure \ref{fig:roi}). Despite its uniqueness problem compared to other biometric traits \cite{rida2016human,rida2018palmprint,bakshi2013security,rida2018robust,fei2017enhanced,dehais2019pbci,rida2018feature,rida2014improved,rida2018efficient,rida2019towards}, footwear impressions hold a great and very promising potential in assisting forensic investigations. For instance, in case of multiple attacks in a short time, it would be unlikely that an attacker would discard or change his/her footwear between different crime places \cite{almaadeed2015partial}. It has also been reported by Alexandre \cite{alexandre1996computerized} that approximately $30\%$ of shoe-prints can be retrieved in SoC. A lifted shoe-print from a SoC can potentially be used in two different tasks:

\begin{itemize}
 \item Match it against a database (such as Foster and Freeman Ltd) in order de determine its model.
 \item Match it against other shoe-prints taken from other SoC to verify if the same shoe model has been used. 
 \end{itemize}

Unfortunately, carrying the matching based on the human knowledge (manually through a paper catalogue or semi-automatically through a computer database) is not a trivial task \cite{kerstholt2007shoe}. Indeed, the limitations of such systems are obvious in case of large databases  due to the need to match the retrieved sample to all database samples (one by one). Furthermore, it is harder to agree on the classification among several users and mostly in case of degraded shoe mark images. This clearly shows the need to a fully automated shoe-print identification system.

Despite the devoted efforts in order to introduce efficient automated computer systems able to search and match shoe-prints, to the best of our knowledge, there is no existing survey bringing together all existing works. The main aim of this paper is to propose a comprehensive overview of existing automatic shoe-print identification. This is intended to provide researchers with state-of-the-art approaches in order to help advance the research topic as well as guiding the neophyte.  Section \ref{arch} presents the main architecture of an automated shoe-print identification system. Section \ref{hol} introduces the holistic techniques. Section \ref{local} describes the local techniques. Section \ref{evaluation} reports the evaluation and obtained performances. Section \ref{discussion} gives the discussion. Finally, Section \ref{conclusion} offers our conclusion.

\section{Automated shoe-print identification }
\label{arch}

\begin{figure}[h]
	\centering
         \includegraphics[width=9cm]{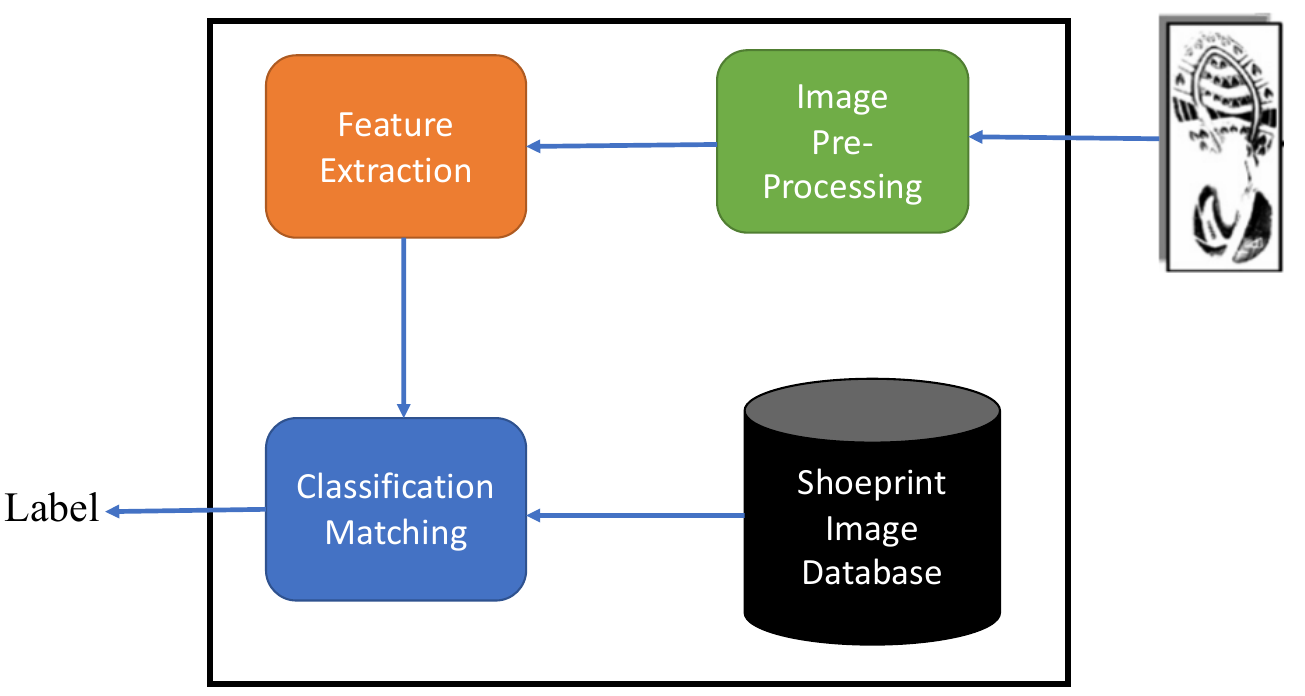}
	\caption{Cohesive schema of the typical processing chain of an automated shoe-print identification system.}
	\label{overview}
\end{figure}

The main architecture of an automated shoe-print identification system can be divided into three main tasks \cite{rida2018comprehensive}: removing the different distortions and enhancing the quality of images by pre-processing, generating discriminative features of a shoe-print using feature extraction techniques and finally classifying/matching the query sample with the whole  database containing the shoe-print models and assigning its class label (i.e. shoe type) using the extracted features and a trained classifier or matching function (see Figure \ref {overview}).

Relevant and discriminative features are of critical and fundamental importance to achieve high performances in any automatic identification system \cite{rida2016robust}. Feature extraction seeks to transform and fix the dimensionality of an initial input raw shoe-print image to generate a new set of features containing meaningful information contributing to assign the observations to the correct corresponding either on training samples or new unseen data class \cite{rida2018ensemble}. Existing state-of-the-art techniques mainly differ by the type of the extracted features. They essentially can be organized in two main categories: holistic and structural methods.


\section{Holistic techniques}
\label{hol}
The holistic or global methods seek to process shoe-print image as a whole. In this context,  Bouridane \textit{et al.} \cite{bouridane2000application} employed Fractal decomposition in order to produce an ensemble of spatial transformations which can reproduce the same image when recursively applied to a nearly similar image. The matching is carried out using Mean Square Noise Error method (MSNE). De Chazal \textit{et al.} \cite{de2005automated} took as features the squared magnitude of the 2D Discrete Fourier Transform (DFT) namely Power Spectral Density (PSD). A 2D correlation function has been used as a  similarly measure and the query image is identified as the one with the highest correlation value in the database.  Based on Oppenheim and Lim \cite{oppenheim1981importance,rida2016gait} assumption claiming that in Fourier domain the phase information is much more important than magnitude in describing the patterns structure, Gueham \textit{et al.} \cite{gueham2007automatic} introduced a Modified Phase-Only Correlation (MPOC) technique through a band pass spectral weighting function. The query sample is then classified as the one with highest matching score. Gueham \textit{et al.} \cite{gueham2008automatic} evaluated two different advanced correlation filters: Optimal Trade-off Synthetic Discriminant Function (OTSDF) and Unconstrained OTSDF. The matching was  carried out using three different metrics, peak height, peak to correlation energy and finally peak to sidelobe ratio. Gueham \textit{et al.} \cite{gueham2008automaticc} exploited Fourier-Mellin transform features obtained by a log-polar mapping followed by a DFT. The matching is performed based on a two dimensional correlation function. AlGarni and Hamiane \cite{algarni2008novel} extracted Hu's moment invariants features, and then four different metrics have been used for the similarity measurement including Euclidean, city block, canberra and correlation. Jing \textit{et al.} \cite{jing2009novel} enhanced the quality of the shoe marks by a pre-processing step including grayscale transformation, noise removal and principal component transformation. Then, four different type of features related to the directionality have been extracted, namely co-occurrence matrices, global Fourier transform, local Fourier transform and directional matrix. Finally, the sum of absolute difference between the previously mentioned features is used as a similarity metric. Patil and Kulkarni \cite{patil2009rotation} have exploited multiresolution features using Gabor transform. In order to be invariant to rotation, Radon transform has been used to estimate the rotation of the shoe-print to compensate the direction of the extracted features. The classification of a new shoe mark image was carried out using nearest-neighbor based on the Euclidean distance. Pei \textit{et al.} \cite{pei2009multiscale} combined odd and even Gabor features to describe the texture and geometry characteristics. Tang and Dai \cite{tang2010automatic} extracted several texture features including the dot texture and shape of edge. Li  \textit{et al.}  \cite{li2014retrieval} combined the integral histogram of the Gabor features with the Euclidean distance and histogram intersection for the similarity measurement. Wei and Gwo \cite{wei2014alignment} used Zernike moments as features and carried out the classification through nearest-neighbor of Euclidean distance. Kong \textit{et al.}  \cite{kong2014novel} extracted  Gabor and Zernike features combined with normalized correlation for matching. Recently and with the progress in machine learning techniques, several learning-based techniques have been proposed, Kortylewski and Vetter \cite{kortylewski2016probabilisticc} suggested a probabilistic compositional active basis model. In the same context, Kong \textit{et al.} \cite{kong2017cross} introduced a multi-channel normalized cross-correlation to match multi-channel deep features extracted by pre-trained convolutional neural network. Wang \textit{et al.} \cite{wang2017manifold} proposed a manifold ranking based method using various extracted features. Recently, Zhang \textit{et al.} \cite{zhang2017adapting} used a pre-trained VGG16 network further tuned using a data augmentation technique.

\begin{table*}[!h]
  \centering
  \caption {Overview of shoe-print identification techniques (features and matching).}
  \noindent
\resizebox{1\linewidth}{!}{
  \begin{tabular}{lll}
    \multicolumn{3}{c}{}  \\
        \toprule
         \toprule
    Techniques & Features &   Classification / Matching  \\
    \midrule
     \midrule
    \tabitem (Bouridane \textit{et al.}, 2000) \cite{bouridane2000application} & Fractal Decomposition &  Mean Square Noise Error    \\
     \tabitem (De Chazal \textit{et al.}., 2005) \cite{de2005automated} & Power Spectral Density  & 2D Correlation   \\
     \tabitem (Gueham \textit{et al.}, 2007) \cite{gueham2007automatic} & Phase & Modified Phase-Only Correlation \\
    \tabitem (Gueham \textit{et al.}, 2008) \cite{gueham2008automatic} & OTSDF+UOTSDF & Peak Height, Peak to Correlation Energy, Peak to Sidelobe Ratio \\
    \tabitem (Gueham \textit{et al.}, 2008) \cite{gueham2008automaticc} & Fourier-Mellin Transform & 2D Correlation \\
    \tabitem (AlGarni and Hamiane, 2008) \cite{algarni2008novel} & Hu's Moments & Euclidean, City-Block, Canberra, Correlation \\
    \tabitem (Jing \textit{et al.}, 2009) \cite{jing2009novel} & Co-occurrence, Global/Local Fourier & Sum of Absolute Difference \\
    \tabitem (Patil and Kulkarni, 2009) \cite{patil2009rotation} & Gabor & Euclidean \\
    \tabitem (Pei \textit{et al.}, 2009) \cite{pei2009multiscale} & Odd and Even Gabor & Tree Similarity \\
    \tabitem (Tang and Dai, 2010) \cite{tang2010automatic} & Texture & Defined Similarity Function \\
    \tabitem (Li \textit{et al.}, 2014) \cite{li2014retrieval} & Gabor  & Euclidean \\
    \tabitem(Wei and Gwo, 2014)  \cite{wei2014alignment} & Zernike Moments & Euclidean \\
    \tabitem (Wei and Gwo, 2014) \cite{wei2014alignment} & Gabor+Zernike & Normalized Correlation \\
    \tabitem (Kortylewski and Vetter, 2016) \cite{kortylewski2016probabilisticc} & Raw Pixels & Probabilistic Model  \\
    \tabitem (Kong \textit{et al.}, 2017) \cite{kong2017cross} & Deep Features & Normalized Cross-Correlation \\
    \tabitem (Wang \textit{et al.}, 2017) \cite{wang2017manifold} & Hybrid Features (Region \& Appearance)  & Manifold Ranking  \\
    \tabitem (Zhang \textit{et al.}, 2017) \cite{zhang2017adapting} & Deep Features & Deep Neural Network \\
    \hline
    \hline
    \tabitem  (Zhang and Allinson, 2005) \cite{zhang2005automatic} &  DFT Histogram Edge Direction& Euclidean \\
     \tabitem (Pavlou and Allinson, 2006) \cite{pavlou2006automatic} & MSER+GLOH+SIFT & Gaussian Weighted Function  \\
     \tabitem (Ghouti \textit{et al.}, 2006) \cite{ghouti2006classification} & Directional FilterBanks &  Euclidean \\
    \tabitem (Su \textit{et al.}, 2007) \cite{su2007local} & MHL+SIFT & Defined Similarity Function \\
    \tabitem (Ramakrishnan and Srihari, 2008) \cite{ramakrishnan2008extraction} & Cosine Similarity+Entropy+Standard Deviation                    &     Conditional Random Fields                     \\
    \tabitem (Pavlou and Allinson, 2009) \cite{pavlou2009automated} & MSER+SIFT & Constraint Kernel  \\
    \tabitem (Nibouche \textit{et al.}, 2009) \cite{nibouche2009rotation} & Multi-Scale Harris+SIFT & RANSAC \\
    \tabitem (Dardi \textit{et al.}, 2009) \cite{dardi2009automatic,dardi2009texture,dardi2009combined} & PSD Mahalanobis Distance & Correlation \\
    \tabitem (Tang \textit{et al.}, 2010) \cite{tang2010footwear} & ISHT+MRHT & Footwear Print Distance \\
    \tabitem (Li \textit{et al.}, 2011) \cite{li2011research}. & SIFT & Cross-Correlation  \\
    \tabitem  (Rathinavel and Arumugam, 2011) \cite{rathinavel2011full}   &         Discrete Cosine Transform & Euclidean \\
    \tabitem  (Hasegawa and Tabbone, 2012) \cite{hasegawa2012local} & HRT & Mean Local Similarity \\
    \tabitem  (Tang \textit{et al.}, 2010, 2012) \cite{tang2010similarity,tang2012efficient}  & ARG & Footwear Print Distance   \\
    \tabitem (Wei \textit{et al.}., 2013) \cite{wei2013use} &     SIFT             &      Cross-Correlation     \\
    \tabitem  (Wang \textit{et al.}, 2014) \cite{wang2014automatic} & Wavelet-Fourier  & 2D Correlation \\
    \tabitem (Kortylewski \textit{et al.}, 2014) \cite{kortylewski2014unsupervised} & Periodicity  &  Defined Similarity Measure \\
    \tabitem (Almaadeed \textit{et al.}, 2015) \cite{almaadeed2015partial} & Harris+Hessian+SIFT & RANSAC \\
    \tabitem (Alizadeh and Kose, 2017) \cite{alizadeh2017automatic}  & Raw Pixels & Sparse Representation for Classification  \\
     \tabitem (Ma \textit{et al.}, 2019) \cite{ma2019shoe}  & Deep Features & Deep Neural Network  \\
    \bottomrule
     \bottomrule
  \end{tabular}}
  \label{tab:sumgei}
\end{table*}

\section{Local techniques}
\label{local}
The local methods try to extract some discriminative features from local shoe-print regions. This includes keypoints or various overlapping/non-overlapping parts (we refer the reader to \cite{krig2016interest} for technical details of different keypoints detection techniques) . Zhang and Allinson \cite{zhang2005automatic} used DFT of the normalized histogram of edge direction as features and the Euclidean distance as measure of similarity. Pavlou and Allinson \cite{pavlou2006automatic} exploited Maximally Stable Extremal Region (MSER) to detect the points of interest followed by Gradient Location and Orientation Histogram (GLOH) and Scale Invariant Feature Transform (SIFT) as feature descriptors. A Gaussian weighted function has been used as similarity metric. Ghouti \textit{et al.} \cite{ghouti2006classification} extracted the block energy-dominant of Directional
FilterBanks (DFBs). The matching was performed using Euclidean distance. Su \textit{et al.} \cite{su2007local} combined the Modified Harris-Laplace (MHL) detector with the enhanced SIFT descriptor. The classification was carried out through nearest-neighbor. Ramakrishnan and Srihari \cite{ramakrishnan2008extraction} proposed a novel technique through the combination of three different features, cosine similarity, entropy and standard deviation with Conditional Random Fields (CRF). Pavlou and Allinson \cite{pavlou2009automated} located points of interest using MSER detector and then the corresponding features are extracted using SIFT descriptor further transformed to an histogram representation. The similarity is measured by a constraint kernel. Nibouche \textit{et al.} \cite{nibouche2009rotation} detected local points of interest through multi-scale Harris detector then SIFT descriptor is applied to extract the features. The matching is carried out iteratively using RANdom SAmple Consensus (RANSAC). Dardi  \textit{et al.}  \cite{dardi2009automatic,dardi2009texture,dardi2009combined} divided the shoe-print image into blocks and then the Mahalanobis distance between all possible block pairs is calculated. The PSD of the obtained distance matrix is used as descriptor and the correlation as similarity measure. Tang \textit{et al.} \cite{tang2010footwear} exploited Iterative Straight-line Hough Transform (ISHT) and Modified Randomized Hough Transform (MRHT). Li \textit{et al.} \cite{li2011research} combined SIFT detector with cross-correlation for matching. Hasegawa and Tabbone \cite{tang2010similarity,hasegawa2012local} decomposed the shoe-print image into connected components and then Histogram Radon Transform (HRT) is used as descriptor to extract the features. The similarity is measured by the mean of local similarities. Rathinavel and Arumugam \cite{rathinavel2011full} extracted Discrete Cosine Transform (DCT) coefficients of overlapped blocks further combined with Principal Component Analysis (PCA) and Fisher Linear Discriminant (FLD). The classification was carried out using nearest-neighbor of Euclidean distance. Tang \textit{et al.} \cite{tang2012efficient} encoded the structural features of shoe-print as an Attributed Relational Graph (ARG) and achieved the matching using a suggested Footwear Print Distance (FPD). Wei \textit{et al.} \cite{wei2013use} combined SIFT features with cross-correlation matching. Wang \textit{et al.} \cite{wang2014automatic} exploited Wavelet-Fourier transform features. Kortylewski \textit{et al.} \cite{kortylewski2014unsupervised} extracted the pattern periodicity features. Almaadeed \cite{almaadeed2015partial} \textit{et al.} combined Harris and Hessian point of interest detectors with SIFT descriptors. The matching is carried out using RANSAC. Alizadeh and Kose \cite{alizadeh2017automatic} proposed an interesting method based on blocked sparse representation. Recently,  Ma \textit{et al.} \cite{ma2019shoe} introduced a novel Multi-Part weighted Convolutional Neural Network (MP-CNN). Table \ref{tab:sumgei} summarizes all the previously mentioned holistic and local shoe-print identification techniques.

\section{Evaluation}
\label{evaluation}
The availability of large and public datasets is essential for a comparative study of the performances including a consistent evaluation. The main noted problem in the research topic of shoe-print identification is the lack or let even say the absence of public benchmarks with pre-defined and standardized evaluation protocols. Most published techniques in the literature were evaluated on non realistic and synthetically generated images by adding artificial distortions such as noise and blur \cite{de2005automated,gueham2008automatic,nibouche2009rotation}. Furthermore, the shoe model databases (i.e. training or gallery) were not made available. Thus a direct and fair comparison of the performance with the reported state-of-the-art techniques is unfortunately not possible. It should be also noted that \cite{dardi2009texture,tang2010footwear} have performed their evaluation based on real data which also was not made available.

Recently, we can notice a new introduced shoe-print database which has been made publicly available for algorithms evaluation namely Footwear Impression Database (FID-300)  \footnote{\url{https://fid.dmi.unibas.ch/}} \cite{kortylewski2014unsupervised}. It has been collected in collaboration between German State Criminal Police Offices of Niedersachsen and Bayern and the company Forensity AG. This database contains 1175 gallery and 300 probe shoe-print images. The probe images has been digitized with a scanner after being lifted with a gel foil from the ground.

Despite the fact that different datasets, partitions and protocols have been used in the evaluation of the aforementioned state-of-the-art techniques, we give a general overview of the obtained performances (summarized in Table \ref{tab:res}). The results are reported in the format X\%@Y, where it refers to the cumulative score X at the first Y matches. It can be clearly seen that various performances have been obtained ranging from 27.10\% to 100\%. This clearly shows the need to public datasets with standardized protocols for the algorithms evaluation.

\begin{table*}[!h]
  \centering
  \caption {Performance of the state-of-the-art methods in shoe-print identification.}
  \noindent
\resizebox{0.95\linewidth}{!}{
  \begin{tabular}{llcl}
    \multicolumn{4}{c}{}  \\
        \toprule
         \toprule
    Techniques  &   Accuracy & Database Size  & Studied Distortions \\
    \midrule
     \midrule
    \tabitem  (Bouridane \textit{et al.}., 2000) \cite{bouridane2000application}   &  88.00\%  @1  &  145  &  rotation \& translation  \\
     \tabitem (De Chazal \textit{et al.}, 2005) \cite{de2005automated} &     87.00\%  @5\% &  475 &  rotation \& translation  \\
     \tabitem  (Zhang and Allinson, 2005) \cite{zhang2005automatic}  & 97.70\% @4\%  &  512 &  rotation, noise, scale \& translation  \\
      \tabitem (Pavlou and Allinson, 2006) \cite{pavlou2006automatic}  & 85.00\% @1  &  368  &  rotation \& translation \\
      \tabitem (Gueham \textit{et al.}, 2007) \cite{gueham2007automatic}  & 100.00\% @1 &  100 &  partial \& noise   \\
      \tabitem (Gueham \textit{et al.}, 2008a) \cite{gueham2008automatic}   & 95.68\% @1 & 100 & rotation, noise \&  occlusion  \\
      \tabitem (AlGarni and Hamiane, 2008) \cite{algarni2008novel} &   99.40\% @1 & 500  &  rotation \& noise  \\
       \tabitem (Gueham \textit{et al.}, 2008b) \cite{gueham2008automaticc}   & 99.00\% @10 & 500 &  rotation, scale, noise  \& occlusion     \\
       \tabitem (Pavlou and Allinson, 2009) \cite{pavlou2009automated}   & 87.00\% @1  &  374 &  - \\
       \tabitem (Dardi \textit{et al.}, 2009a) \cite{dardi2009automatic}  &  49.00\% @1 &  87  &  noise  \\
        \tabitem (Nibouche \textit{et al.}, 2009) \cite{nibouche2009rotation}  & 90.00\% @1  &  300  &  rotation, noise \& occlusion  \\
        \tabitem (Patil and Kulkarni, 2009) \cite{patil2009rotation}  & 91.00\% @1 & 1400 &  rotation, noise \& occlusion   \\
        \tabitem (Pei \textit{et al.}, 2009) \cite{pei2009multiscale}  & 61.70\% @5  &  6000 &  noise \& occlusion  \\
        \tabitem (Dardi \textit{et al.}, 2009c) \cite{dardi2009texture}  & 73.00\% @10 &  87  &  rotation, scale \& translation  \\
        \tabitem (Tang \textit{et al.}, 2010b) \cite{tang2010footwear}   & 71.00\% @1\%  &  2660  &  rotation, scale, translation \& occlusion  \\
        \tabitem  (Tang \textit{et al.}, 2012) \cite{tang2012efficient}   & 70.00\% @1\%  &  2660  &  rotation, scale, translation \& noise   \\
         \tabitem  (Wang \textit{et al.}, 2014) \cite{wang2014automatic}    & 90.87\% @2\%  &  210 000 &  rotation, translation \&  scale   \\
          \tabitem (Kortylewski \textit{et al.}, 2014) \cite{kortylewski2014unsupervised}    &  27.10\% @1\% &  1175  &  translation \& noise   \\
         \tabitem (Almaadeed \textit{et al.}, 2015) \cite{almaadeed2015partial}  & 99.33\% @1 &  300 &  rotation, scale, noise \& occlusion  \\
          \tabitem (Kortylewski and Vetter, 2016) \cite{kortylewski2016probabilisticc}   & 71.00\% @20\% & 1175 &  -    \\
          \tabitem (Alizadeh and Kose, 2017) \cite{alizadeh2017automatic}   & 99.47\% @1 &  190  &  noise, rotation  \& occlusion\\

    \bottomrule
     \bottomrule
  \end{tabular}}
  \label{tab:res}
\end{table*}

\begin{figure}[h]
	\centering
         \includegraphics[width=5cm]{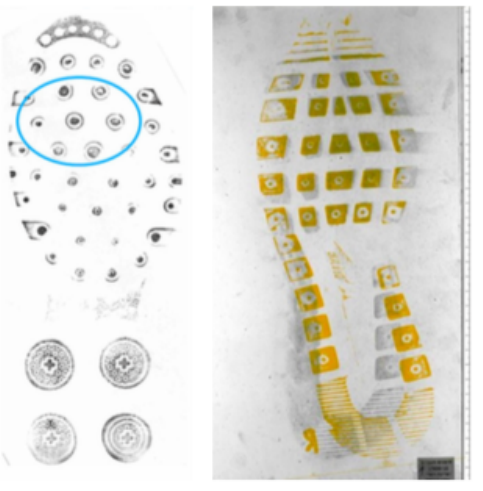}
	\caption{Non-rigid deformation between probe image (left) and its gallery image (right) \cite{kortylewski2017model}.    Blue circle stands for deformation.}
	\label{ove}
\end{figure}
\begin{figure}[h]
	\centering
         \includegraphics[width=2.8cm]{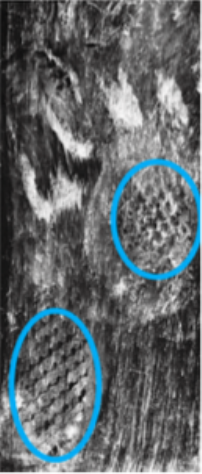}
	\caption{Shoe-print with structured background \cite{kortylewski2017model}.Blue circles stand for shoe-prints.}
	\label{background}
\end{figure}

Actually, automatic shoe-print identification is a very challenging task in computer vision systems. Indeed, it suffers from different variations in shape and appearance due to the tread material and properties of surface (Figure \ref{ove}). Furthermore, shoe-prints are cluttered since gallery images have no background while probe ones have a complicated and structured background which is hardly distinguishable from patterns of interest (Figure \ref{background}). In addition to that, occlusion, noise, translation and limited training data are further problems \cite{kortylewski2017model}.

 \section{Discussion and Current Challenges}
 \label{discussion}
 
A considerable amount of techniques have been introduced in order to the tackle the problem of shoe-print identification using a large variety of features. These extracted features determine which information and properties are available during the identification process \cite{rida2017improved}. They should capture enough invariant properties within the same shoe class and variant ones between different ones \cite{rida2015unsupervised,micheletto2018multiple}. The conventional methods to identify the lifted shoe marks are mainly based on low-level handcrafted features designed based on the human knowledge. Unfortunately, despite their good performances in some controlled and specific tasks, handcrafted representations are usually ad-hoc, tend to overfitting and lack of generalization ability in various realistic scenarios. Indeed,  shoe-print identification is not trial task due to the large intra-class variations caused by the rotation, noise, occlusion, translation and scale distortions. This clearly shows the need to robust techniques capable to operate in complicated and degraded scenarios.

In contrast to handcrafted feature engineering, feature learning approaches are capable to learn robust, discriminative and data-driven representations from the raw data without making use of any prior knowledge of the task \cite{al2018palmprint,rida2018palmprintt,rida2018novel}. Among the involved techniques we can find deep learning with the goal of end-to-end identification system \cite{lecun2015deep}. It seeks to stack more than the usual two neural layers where each layer encodes some specific properties further combined in order to learn representative and discriminative representations. Among the existing deep learning models which can potentially be applied to shoe-print identification, we can find Convolutional Neural Networks (CNN) \cite{lecun1998gradient}. They seek to learn discriminative representations with invariant properties.

Up to day, handcrafted feature represent the most and widely used features for shoe mark identification since the deep models require a considerable and huge amount of data in order to be reliable. Unfortunately, the existing shoe-print identification datasets have a very limited size and mainly one example per each shoe class. To be effective and tackle the problem of limited training data, a possible solution is transfer learning. It consists in exploiting models that have been already pre-trained on a huge amount of data for another task followed by a fine tuning step to fit the model to the target application.

 \section{Conclusion}
\label{conclusion}

shoe-print represents an important clue in scene crime for the proper progress of investigations in order identify the criminals. A large variety of handcrafted features have been used for automatic shoe-print identification. These features have shown good performance in limited and controlled scenarios. Unfortunately, they fail when they are dealing with large intra-class variations caused by the noise, occlusions, rotation and various scale distortions. A good alternative to these conventional features are the learned ones, e.g. deep learning, which have more  generalization ability in more complicated scenarios. To be effective, these models need to be trained on a large amount of data.

Large and public datasets are essential and of extreme importance for any comparative study of the performances including a consistent evaluation. The main noted problem in the research topic of shoe-print identification is the absence of public benchmarks with pre-defined and standardized evaluation protocols. Most published techniques in the literature were evaluated on non realistic and synthetically generated images. This is clearly show the need to build new large datasets in order to boost the shoe-print research topic.


\begin{thebibliography}{10}
\expandafter\ifx\csname url\endcsname\relax
  \def\url#1{\texttt{#1}}\fi
\expandafter\ifx\csname urlprefix\endcsname\relax\def\urlprefix{URL }\fi
\expandafter\ifx\csname href\endcsname\relax
  \def\href#1#2{#2} \def\path#1{#1}\fi

\bibitem{huynh2003automatic}
C.~Huynh, P.~de~Chazal, D.~McErlean, R.~B. Reilly, T.~J. Hannigan, L.~M.
  Fleury, Automatic classification of shoeprints for use in forensic science
  based on the fourier transform, in: IEEE International Conference on Image
  Processing (ICIP), 2003, Vol.~3, 2003, pp. 569--572, {DOI:}
  10.1109/ICIP.2003.1247308.

\bibitem{locard1930analysis}
E.~Locard, The analysis of dust traces, Am. J. Police Sci. 1 (1930) 276, {DOI:}
  10.1007/s001140050.

\bibitem{vagavc2017detection}
M.~Vaga{\v{c}}, M.~Povinsk{\`y}, M.~Melicher{\v{c}}{\'\i}k, Detection of shoe
  sole features using dnn, in: 14th IEEE International Scientific Conference on
  Informatics, 2017, 2017, pp. 416--419, {DOI:}
  10.1109/INFORMATICS.2017.8327285.

\bibitem{liu2017multi}
Y.~Liu, D.~Hu, J.~Fan, F.~Wang, D.~Zhang, Multi-feature fusion for crime scene
  investigation image retrieval, in: IEEE International Conference on Digital
  Image Computing: Techniques and Applications (DICTA), 2017, 2017, pp. 1--7,
  {DOI:} 10.1109/DICTA.2017.8227466.

\bibitem{benecke1997dna}
M.~Benecke, Dna typing in forensic medicine and in criminal investigations: a
  current survey, Naturwissenschaften 84~(5) (1997) 181--188, {DOI:}
  10.1007/s001140050375.

\bibitem{robertson2002forensic}
J.~R. Robertson, Forensic examination of hair, CRC Press, 2002, {DOI:}
  10.1080/00450610109410825.

\bibitem{buckleton2016forensic}
J.~S. Buckleton, J.-A. Bright, D.~Taylor, Forensic DNA evidence interpretation,
  CRC press, 2016, {ISBN:} 9781482258899.

\bibitem{robertson2016interpreting}
B.~Robertson, G.~A. Vignaux, C.~E. Berger, Interpreting evidence: evaluating
  forensic science in the courtroom, John Wiley \& Sons, 2016, {ISBN:}
  978-1-118-49243-7.

\bibitem{rida2015improved}
I.~Rida, A.~Bouridane, G.~L. Marcialis, P.~Tuveri, Improved human gait
  recognition, in: International Conference on Image Analysis and Processing,
  Springer, 2015, pp. 119--129, {DOI:} 10.1007/978-3-319-23234-8\_12.

\bibitem{thompson2006forensic}
T.~Thompson, S.~Black, Forensic human identification: An introduction, CRC
  press, 2006, {ISBN:} 9780849339547.

\bibitem{bodziak2017footwear}
W.~J. Bodziak, Footwear impression evidence: detection, recovery and
  examination, CRC Press, 2017, {ISBN:} 9780849310454.

\bibitem{kong2017cross}
B.~Kong, D.~Ramanan, C.~Fowlkes, Cross-domain forensic shoeprint matching, in:
  British Machine Vision Conference (BMVC), 2017, pp. 1--5, {DOI:}
  10.5244/C.21.38.

\bibitem{rida2016human}
I.~Rida, X.~Jiang, G.~L. Marcialis, Human body part selection by group lasso of
  motion for model-free gait recognition, IEEE Signal Processing Letters 23~(1)
  (2016) 154--158, {DOI:} 10.1109/LSP.2015.2507200.

\bibitem{rida2018palmprint}
I.~Rida, R.~Herault, G.~L. Marcialis, G.~Gasso, Palmprint recognition with an
  efficient data driven ensemble classifier, Pattern Recognition Letters{DOI:}
  10.1016/j.patrec.2018.04.033.

\bibitem{bakshi2013security}
S.~Bakshi, T.~Tuglular, Security through human-factors and biometrics, in: 6th
  International Conference on Security of Information and Networks, ACM, 2013,
  pp. 463--463, {DOI:} 10.1145/2523514.2523597.

\bibitem{rida2018robust}
I.~Rida, N.~Al-maadeed, S.~Al-maadeed, Robust gait recognition: a comprehensive
  survey, IET Biometrics{DOI:} 10.1049/iet-bmt.2018.5063.

\bibitem{fei2017enhanced}
L.~Fei, S.~Teng, J.~Wu, I.~Rida, Enhanced minutiae extraction for
  high-resolution palmprint recognition, International Journal of Image and
  Graphics 17~(04) (2017) 1750020, {DOI:} 10.1142/S0219467817500206.

\bibitem{dehais2019pbci}
F.~Dehais, I.~Rida, R.~N. Roy, J.~Iversen, T.~Mullen, D.~Callan, A pbci to
  predict attentional error before it happens in real flight conditions, in:
  2019 IEEE International Conference on Systems, Man and Cybernetics (SMC),
  IEEE, 2019, pp. 4155--4160, {DOI:} 10.1109/SMC.2019.8914010.

\bibitem{rida2018feature}
I.~Rida, Feature extraction for temporal signal recognition: An overview, arXiv
  preprint arXiv:1812.01780.

\bibitem{rida2014improved}
I.~Rida, S.~Almaadeed, A.~Bouridane, Improved gait recognition based on gait
  energy images, in: 2014 26th International Conference on Microelectronics
  (ICM), IEEE, 2014, pp. 40--43, {DOI:} 10.1109/ICM.2014.7071801.

\bibitem{rida2018efficient}
I.~Rida, R.~H{\'e}rault, G.~Gasso, An efficient supervised dictionary learning
  method for audio signal recognition, arXiv preprint arXiv:1812.04748.

\bibitem{rida2019towards}
I.~Rida, Towards human body-part learning for model-free gait recognition,
  arXiv preprint arXiv:1904.01620.

\bibitem{almaadeed2015partial}
S.~Almaadeed, A.~Bouridane, D.~Crookes, O.~Nibouche, Partial shoeprint
  retrieval using multiple point-of-interest detectors and sift descriptors,
  Integrated Computer-Aided Engineering 22~(1) (2015) 41--58, {DOI:}
  10.3233/ICA-140480.

\bibitem{alexandre1996computerized}
G.~Alexandre, Computerized classification of the shoeprints of burglars' soles,
  Forensic Science International 82~(1) (1996) 59--65, {DOI:}
  10.1016/0379-0738(96)01967-6.

\bibitem{kerstholt2007shoe}
J.~H. Kerstholt, R.~Paashuis, M.~Sjerps, Shoe print examinations: effects of
  expectation, complexity and experience, Forensic science international
  165~(1) (2007) 30--34, {DOI:} 10.1016/j.forsciint.2006.02.039.

\bibitem{rida2018comprehensive}
I.~Rida, N.~Al-Maadeed, S.~Al-Maadeed, S.~Bakshi, A comprehensive overview of
  feature representation for biometric recognition, Multimedia Tools and
  Applications (2018) 1--24{DOI:} 10.1007/s11042-018-6808-5.

\bibitem{rida2016robust}
I.~Rida, L.~Boubchir, N.~Al-Maadeed, S.~Al-Maadeed, A.~Bouridane, Robust
  model-free gait recognition by statistical dependency feature selection and
  globality-locality preserving projections, in: 39th IEEE International
  Conference on Telecommunications and Signal Processing (TSP), 2016, 2016, pp.
  652--655, {DOI:} 10.1109/TSP.2016.7760963.

\bibitem{rida2018ensemble}
I.~Rida, S.~Al~Maadeed, X.~Jiang, F.~Lunke, A.~Bensrhair, An ensemble learning
  method based on random subspace sampling for palmprint identification, in:
  2018 IEEE International conference on acoustics, speech and signal processing
  (ICASSP), IEEE, 2018, pp. 2047--2051, {DOI:} 10.1109/ICASSP.2018.8462051.

\bibitem{bouridane2000application}
A.~Bouridane, A.~Alexander, M.~Nibouche, D.~Crookes, Application of fractals to
  the detection and classification of shoeprints, in: IEEE International
  Conference on Image Processing (ICIP), 2000, Vol.~1, 2000, pp. 474--477,
  {DOI:} 10.1109/ICIP.2000.900998.

\bibitem{de2005automated}
P.~De~Chazal, J.~Flynn, R.~B. Reilly, Automated processing of shoeprint images
  based on the fourier transform for use in forensic science, IEEE transactions
  on pattern analysis and machine intelligence 27~(3) (2005) 341--350, {DOI:}
  10.1109/TPAMI.2005.48.

\bibitem{oppenheim1981importance}
A.~V. Oppenheim, J.~S. Lim, The importance of phase in signals, Proceedings of
  the IEEE 69~(5) (1981) 529--541, {DOI:} 10.1109/PROC.1981.12022.

\bibitem{rida2016gait}
I.~Rida, S.~Almaadeed, A.~Bouridane, Gait recognition based on modified
  phase-only correlation, Signal, Image and Video Processing 10~(3) (2016)
  463--470, {DOI:} 10.1007/s11760-015-0766-4.

\bibitem{gueham2007automatic}
M.~Gueham, A.~Bouridane, D.~Crookes, Automatic recognition of partial
  shoeprints based on phase-only correlation, in: IEEE International Conference
  on Image Processing (ICIP), 2007, Vol.~4, 2007, pp. 441--444, {DOI:}
  10.1109/ICIP.2007.4380049.

\bibitem{gueham2008automatic}
M.~Gueham, A.~Bouridane, D.~Crookes, Automatic classification of partial
  shoeprints using advanced correlation filters for use in forensic science,
  in: 19th IEEE International Conference on Pattern Recognition (ICPR), 2008,
  2008, pp. 1--4, {DOI:} 10.1109/ICPR.2008.4761058.

\bibitem{gueham2008automaticc}
M.~Gueham, A.~Bouridane, D.~Crookes, O.~Nibouche, Automatic recognition of
  shoeprints using fourier-mellin transform, in: IEEE Conference on Adaptive
  Hardware and Systems (AHS), 2008, 2008, pp. 487--491, {DOI:}
  10.1109/AHS.2008.48.

\bibitem{algarni2008novel}
G.~AlGarni, M.~Hamiane, A novel technique for automatic shoeprint image
  retrieval, Forensic science international 181~(1-3) (2008) 10--14, {DOI:}
  10.1016/j.forsciint.2008.07.004.

\bibitem{jing2009novel}
M.-Q. Jing, W.-J. Ho, L.-H. Chen, A novel method for shoeprints recognition and
  classification, in: IEEE International Conference on Machine Learning and
  Cybernetics, 2009, Vol.~5, 2009, pp. 2846--2851, {DOI:}
  10.1109/ICMLC.2009.5212580.

\bibitem{patil2009rotation}
P.~M. Patil, J.~V. Kulkarni, Rotation and intensity invariant shoeprint
  matching using gabor transform with application to forensic science, Pattern
  Recognition 42~(7) (2009) 1308--1317, {DOI:} 10.1016/j.patcog.2008.11.008.

\bibitem{pei2009multiscale}
W.~Pei, Y.-y. Zhu, Y.-n. Na, X.-g. He, Multiscale gabor wavelet for shoeprint
  image retrieval, in: 2nd IEEE International Congress on Image and Signal
  Processing (CISP), 2009, 2009, pp. 1--5, {DOI:} 10.1109/CISP.2009.5304124.

\bibitem{tang2010automatic}
C.~Tang, X.~Dai, Automatic shoe sole pattern retrieval system based on image
  content of shoeprint, in: IEEE International Conference on Computer Design
  and Applications (ICCDA), 2010, Vol.~4, 2010, pp. 602--605, {DOI:}
  10.1109/DICTA.2017.8227466.

\bibitem{li2014retrieval}
X.~Li, M.~Wu, Z.~Shi, The retrieval of shoeprint images based on the integral
  histogram of the gabor transform domain, in: International Conference on
  Intelligent Information Processing, Springer, 2014, pp. 249--258, {DOI:}
  10.1504/IJGCRSIS.2012.049981.

\bibitem{wei2014alignment}
C.-H. Wei, C.-Y. Gwo, Alignment of core point for shoeprint analysis and
  retrieval, in: IEEE International Conference on Information Science,
  Electronics and Electrical Engineering (ISEEE), 2014, Vol.~2, IEEE, 2014, pp.
  1069--1072, {DOI:} 10.1109/InfoSEEE.2014.6947833.

\bibitem{kong2014novel}
X.~Kong, C.~Yang, F.~Zheng, A novel method for shoeprint recognition in crime
  scenes, in: Chinese Conference on Biometric Recognition, Springer, 2014, pp.
  498--505, {DOI:} 10.1007/978-3-319-12484-1\_57.

\bibitem{kortylewski2016probabilisticc}
A.~Kortylewski, T.~Vetter, Probabilistic compositional active basis models for
  robust pattern recognition., in: British Machine Vision Conference (BMVC),
  2016, pp. 1--12.

\bibitem{wang2017manifold}
X.~Wang, C.~Zhang, Y.~Wu, Y.~Shu, A manifold ranking based method using hybrid
  features for crime scene shoeprint retrieval, Multimedia Tools and
  Applications 76~(20) (2017) 21629--21649, {DOI:} 10.1007/s11042-016-4029-3.

\bibitem{zhang2017adapting}
Y.~Zhang, H.~Fu, E.~Dellandr{\'e}a, L.~Chen, Adapting convolutional neural
  networks on the shoeprint retrieval for forensic use, in: Chinese Conference
  on Biometric Recognition, Springer, 2017, pp. 520--527, {DOI:}
  10.1109/InfoSEEE.2014.6947833.

\bibitem{zhang2005automatic}
L.~Zhang, N.~Allinson, Automatic shoeprint retrieval system for use in forensic
  investigations, in: UK Workshop On Computational Intelligence, Vol.~99, 2005,
  pp. 137--142.

\bibitem{pavlou2006automatic}
M.~Pavlou, N.~M. Allinson, Automatic extraction and classification of footwear
  patterns, in: International Conference on Intelligent Data Engineering and
  Automated Learning, Springer, 2006, pp. 721--728, {DOI:}
  10.1007/11875581\_87.

\bibitem{ghouti2006classification}
L.~Ghouti, A.~Bouridane, D.~Crookes, Classification of shoeprint images using
  directional filter banks, in: International Conference on Visual Information
  Engineering (VIE), 2006, IET, 2006, pp. 167--173, {DOI:} 10.1049/cp:20060522.

\bibitem{su2007local}
H.~Su, D.~Crookes, A.~Bouridane, M.~Gueham, Local image features for shoeprint
  image retrieval, in: British machine vision conference (BMVC), Vol. 2007,
  2007, pp. 1--10, {DOI:} 10.5244/C.21.38.

\bibitem{ramakrishnan2008extraction}
V.~Ramakrishnan, S.~Srihari, Extraction of shoe-print patterns from impression
  evidence using conditional random fields, in: 19th IEEE International
  Conference on Pattern Recognition (ICPR), 2008, 2008, pp. 1--4, {DOI:}
  10.1109/ICPR.2008.4761881.

\bibitem{pavlou2009automated}
M.~Pavlou, N.~M. Allinson, Automated encoding of footwear patterns for fast
  indexing, Image and Vision Computing 27~(4) (2009) 402--409, {DOI:}
  10.1016/j.imavis.2008.06.003.

\bibitem{nibouche2009rotation}
O.~Nibouche, A.~Bouridane, M.~Gueham, M.~Laadjel, Rotation invariant matching
  of partial shoeprints, in: 13th IEEE International Machine Vision and Image
  Processing (IMVIP), 2009, 2009, pp. 94--98, {DOI:} 10.1109/IMVIP.2009.24.

\bibitem{dardi2009automatic}
F.~Dardi, F.~Cervelli, S.~Carrato, An automatic footwear retrieval system for
  shoe marks from real crime scenes, in: 6th IEEE International Symposium on
  Image and Signal Processing and Analysis (ISPA), 2009, 2009, pp. 668--672,
  {DOI:} 10.1109/ISPA.2009.5297667.

\bibitem{dardi2009texture}
F.~Dardi, F.~Cervelli, S.~Carrato, A texture based shoe retrieval system for
  shoe marks of real crime scenes, in: International Conference on Image
  Analysis and Processing (ICIAP), Springer, 2009, pp. 384--393, {DOI:}
  10.1007/978-3-642-04146-4\_42.

\bibitem{dardi2009combined}
F.~Dardi, F.~Cervelli, S.~Carrato, A combined approach for footwear retrieval
  of crime scene shoe marks, in: 3rd International Conference on Crime
  Detection and Prevention (ICDP), 2009, IET, 2009, pp. 1--6, {DOI:}
  10.1049/ic.2009.0237.

\bibitem{tang2010footwear}
Y.~Tang, S.~N. Srihari, H.~Kasiviswanathan, J.~J. Corso, Footwear print
  retrieval system for real crime scene marks, in: International Workshop on
  Computational Forensics, Springer, 2010, pp. 88--100, {DOI:}
  10.1007/978-3-642-19376-7\_8.

\bibitem{li2011research}
Z.~Li, C.~Wei, Y.~Li, T.~Sun, Research of shoeprint image stream retrival
  algorithm with scale-invariance feature transform, in: IEEE International
  Conference on Multimedia Technology (ICMT), 2011, 2011, pp. 5488--5491,
  {DOI:} 10.1109/ICMT.2011.6002147.

\bibitem{rathinavel2011full}
S.~Rathinavel, S.~Arumugam, Full shoe print recognition based on pass band dct
  and partial shoe print identification using overlapped block method for
  degraded images, International Journal of Computer Applications 26~(8) (2011)
  16--21, {DOI:} 10.5120/3126-4301.

\bibitem{hasegawa2012local}
M.~Hasegawa, S.~Tabbone, A local adaptation of the histogram radon transform
  descriptor: an application to a shoe print dataset, in: Joint IAPR
  International Workshops on Statistical Techniques in Pattern Recognition
  (SPR) and Structural and Syntactic Pattern Recognition (SSPR), Springer,
  2012, pp. 675--683, {DOI:} 10.1007/978-3-642-34166-3\_74.

\bibitem{tang2010similarity}
Y.~Tang, S.~N. Srihari, H.~Kasiviswanathan, Similarity and clustering of
  footwear prints, in: IEEE International Conference on Granular Computing
  (GrC), 2010, 2010, pp. 459--464, {DOI:} 10.1109/GrC.2010.175.

\bibitem{tang2012efficient}
Y.~Tang, H.~Kasiviswanathan, S.~N. Srihari, An efficient clustering-based
  retrieval framework for real crime scene footwear marks, International
  Journal of Granular Computing, Rough Sets and Intelligent Systems 2~(4)
  (2012) 327--360, {DOI:} 10.1504/IJGCRSIS.2012.049981.

\bibitem{wei2013use}
C.-H. Wei, Y.~Li, C.-Y. Gwo, The use of scale-invariance feature transform
  approach to recognize and retrieve incomplete shoeprints, Journal of forensic
  sciences 58~(3) (2013) 625--630, {DOI:} 10.1111/1556-4029.12089.

\bibitem{wang2014automatic}
X.~Wang, H.~Sun, Q.~Yu, C.~Zhang, Automatic shoeprint retrieval algorithm for
  real crime scenes, in: Asian Conference on Computer Vision (ACCV), Springer,
  2014, pp. 399--413, {DOI:} 10.1016/j.forsciint.2017.05.025.

\bibitem{kortylewski2014unsupervised}
A.~Kortylewski, T.~Albrecht, T.~Vetter, Unsupervised footwear impression
  analysis and retrieval from crime scene data, in: Asian Conference on
  Computer Vision (ACCV), Springer, 2014, pp. 644--658, {DOI:}
  10.1007/978-3-319-16628-5\_46.

\bibitem{alizadeh2017automatic}
S.~Alizadeh, C.~Kose, Automatic retrieval of shoeprint images using blocked
  sparse representation, Forensic science international 277 (2017) 103--114,
  {DOI:} 10.1016/j.forsciint.2017.05.025.

\bibitem{ma2019shoe}
Z.~Ma, Y.~Ding, S.~Wen, J.~Xie, Y.~Jin, Z.~Si, H.~Wang, Shoe-print image
  retrieval with multi-part weighted cnn, IEEE Access 7 (2019) 59728--59736.

\bibitem{krig2016interest}
S.~Krig, Interest point detector and feature descriptor survey, in: Computer
  Vision Metrics, Springer, 2016, pp. 187--246, {DOI:}
  10.1007/978-3-319-33762-3\_6.

\bibitem{kortylewski2017model}
A.~Kortylewski, Model-based image analysis for forensic shoe print recognition,
  Ph.D. thesis, University\_of\_Basel (2017).

\bibitem{rida2017improved}
I.~Rida, N.~Al~Maadeed, G.~L. Marcialis, A.~Bouridane, R.~Herault, G.~Gasso,
  Improved model-free gait recognition based on human body part, in: Biometric
  Security and Privacy, Springer, 2017, pp. 141--161, {DOI:}
  10.1007/978-3-319-47301-7\_6.

\bibitem{rida2015unsupervised}
I.~Rida, S.~Al~Maadeed, A.~Bouridane, Unsupervised feature selection method for
  improved human gait recognition, in: Signal Processing Conference (EUSIPCO),
  2015 23rd European, IEEE, 2015, pp. 1128--1132, {DOI:}
  10.1109/EUSIPCO.2015.7362559.

\bibitem{micheletto2018multiple}
M.~Micheletto, G.~Orr{\`u}, I.~Rida, L.~Ghiani, G.~L. Marcialis, A multiple
  classifiers-based approach to palmvein identification, in: 2018 Eighth
  International Conference on Image Processing Theory, Tools and Applications
  (IPTA), IEEE, 2018, pp. 1--6, {DOI:} 10.1109/IPTA.2018.8608150.

\bibitem{al2018palmprint}
S.~Al~Maadeed, X.~Jiang, I.~Rida, A.~Bouridane, Palmprint identification using
  sparse and dense hybrid representation, Multimedia Tools and Applications
  (2018) 1--15{DOI:} 10.1007/s11042-018-5655-8.

\bibitem{rida2018palmprintt}
I.~Rida, S.~Al-Maadeed, A.~Mahmood, A.~Bouridane, S.~Bakshi, Palmprint
  identification using an ensemble of sparse representations, IEEE Access 6
  (2018) 3241--3248, {DOI:} 10.1109/ACCESS.2017.2787666.

\bibitem{rida2018novel}
I.~Rida, N.~Al~Maadeed, S.~Al~Maadeed, A novel efficient classwise sparse and
  collaborative representation for holistic palmprint recognition, in: 2018
  NASA/ESA Conference on Adaptive Hardware and Systems (AHS), IEEE, 2018, pp.
  156--161, {DOI:} 10.1109/AHS.2018.8541428.

\bibitem{lecun2015deep}
Y.~LeCun, Y.~Bengio, G.~Hinton, Deep learning, nature 521~(7553) (2015) 436,
  {DOI:} 10.1038/nature14539.

\bibitem{lecun1998gradient}
Y.~LeCun, L.~Bottou, Y.~Bengio, P.~Haffner, Gradient-based learning applied to
  document recognition, Proceedings of the IEEE 86~(11) (1998) 2278--2324,
  {DOI:} 10.1109/5.726791.

\end{thebibliography}
\end{document}